\documentclass[twocolumn]{article}
\usepackage{absspconf,amsmath,epsfig}
\usepackage{subcaption}
\usepackage{booktabs}
\usepackage{siunitx}

\usepackage[dvipsnames]{xcolor}
\title{Toward Foundation Models for Earth Monitoring:\\ Generalizable Deep Learning Models for Natural Hazard Segmentation}
\name{Johannes Jakubik$^{1,2}\sthanks{work was done during research internship at IBM Research, Rüschlikon}$, Michal Muszynski$^{3}$, Michael Vössing$^{1,2}$, Niklas Kühl$^{1,4}$, Thomas Brunschwiler$^{3}$}
\address{$^{1}$Karlsruhe Institute of Technology, $^{2}$IBM Germany, $^{3}$IBM Research Europe, $^{4}$University of Bayreuth}

\begin{document}
\maketitle
\begin{abstract}
Climate change results in an increased probability of extreme weather events that put societies and businesses at risk on a global scale. Therefore, near real-time mapping of natural hazards is an emerging priority for the support of natural disaster relief, risk management, and informed governmental policy decisions. Current remote sensing based approaches to near real-time natural hazard mapping increasingly leverage advantage of deep learning (DL). Nevertheless, DL-based approaches are mainly designed for one specific task in a single geographic region based on specific frequency bands of satellite data. For that reason, DL models used to map specific natural hazards struggle with their generalization to other types of natural hazards in unseen regions. In this work, we propose a methodology to significantly improve the generalizability of DL natural hazards mappers based on pre-training on a suitable pre-task. Without access to any data from the target domain, we demonstrate that this methodology improved generalizability across four U-Net architectures for the segmentation of unseen natural hazards, such as flood events, landslides, and massive glacier collapses. Importantly, our method is strongly invariant to geographic differences and the type of input frequency bands of satellite data. That is confirmed by obtaining a balanced accuracy of up to 0.74 in comparison with performance of reference baselines. By leveraging characteristics of unlabeled images from the target domain that are publicly available, our approach is able to further improve the generalization behavior of DL models without fine-tuning. That is reflected in performance metrics. Thereby, our approach is one of first attempts to support the development of foundation models for earth monitoring with the objective of directly segmenting unseen natural hazards across novel geographic regions from different sources of satellite imagery.
\end{abstract}


\section{Motivation}

Climate change leads to an increased probability of extreme weather events causing corresponding severe floods \cite{hallegatte2013future}, landslides \cite{kainthura2022hybrid}, and massive glacier collapses \cite{rounce2023global} beyond other natural hazards.
That puts critical infrastructures, such as means of transport, energy and water supplies, and telecommunication 
at high risk of failure, causing subsequent extreme weather events, and decreasing the quality of human lives on a global scale.
Accurate and near real-time mapping of natural hazards is emerging priority for support of natural disaster relief, risk management, and informed governmental policy decisions. 
    
Satellite imagery obtained from remote sensing technologies has become one of the most common techniques used to detect and segment extents of natural hazards on a large scale because of its increased availability, reduced costs, and growing global coverage over the few last decades.
Leveraging this data is key to approach a better understanding and response to natural hazards. Importantly, the amount of satellite data is larger than the amount of available natural language processing data \cite{ma2015remote}, which results in a large number of attempts towards foundation models for remote sensing (e.g., \cite{wang2022advancing,sun2022ringmo}.

Advances in Deep Learning (DL) and computer vision have been creating a set of tools for remote sensing of the Earth’s surface \cite{zhu2017deep,wang2022self}. For instance, DL approaches to detect clouds significantly outperform physics-based methods \cite{zantedeschi2019cumulo}. DL models have also shown their advantages for land cover classification tasks \cite{wang2020weakly} and segmentation of floods \cite{fraccaro2022deploying}.
Nevertheless, DL based approaches have yet been designed to detect a given type of natural hazards in a selected geographic region based on specific modalities of satellite data (e.g., specific frequency bands).
For that reason, DL models applied to mapping selected natural hazards struggle with their limited generalization to predict other types of natural hazards.  
Adaption of DL models to unseen natural hazards or novel geographic regions requires additional data labeled by domain experts for fine-tuning of the mapping model. Data annotation process takes a significant amount of time, prolongs model fine-tuning, and generates additional costs.

In this work, we propose a methodology to significantly improve generalizability abilities of DL based natural hazards mappers based on pre-training of DL models on a suitable pre-task in the context of earth observations.
Our contributions are as follows: Firstly, we demonstrate that our approach significantly improves the generalizability of DL models, i.e., an increase of performance of four U-Net architectures applied to the segmentation of unseen natural hazards. Secondly, we show that our approach generalizes over: novel geographic regions of interests and different frequency bands of satellite data without access to target domain data. 
Thirdly, by leveraging characteristics of few unlabeled data (i.e., images without annotations) from the target domain that are publicly available, our approach is able to further improve the generalization properties of DL based mappers. 

\section{Methodology}

To improve the generalization of DL models, we proposed to pre-train well-established U-Net models on a suitable pre-task consisting of high-resolution satellite images. 
The goal of our work is neither to achieve the highest possible performance for selected tasks nor to propose a novel and advanced architecture that could lead to model overfitting to the problem. 
In contrast, the main objective of our work is to prove the concept of generalizing deep learning models to unseen geo-spatial tasks by applying our proposed methodology.
We make use of four well-known convolutional neural network architectures as backbones of the U-Net:
ResNet-34 \cite{he2016deep}, SeNet-154 \cite{hu2018squeeze}, SeResNeXt-50 \cite{hu2018squeeze,xie2017aggregated}, DPN-92 \cite{chen2017dual} that are suitable for the sample size of our source task (i.e., 9,168 pre-disaster 1024x1024 high-resolution color images). 
Since pre-training on well-known natural image datasets does not allow us to generalize the natural hazard mappers to the segmentation of unseen natural hazards in satellite images, we select the localization of buildings based on earth observations as the most relevant pre-task for our study \cite{gupta2019xbd}.

\subsection{Datasets:} 
In the section, we introduce a dataset we utilize for pre-training of U-Net models and three datasets for three different remote sensing downstream tasks.

\noindent  \textbf{xBD dataset:} 
We pre-trained all models on the building localization task from the xBD dataset \cite{gupta2019xbd}. The dataset consists of labeled, high-resolution satellite images covering over 45,000 square km across diverse geographic regions, where buildings were damaged by weather hazards (i.e., wildfires, earthquakes, and hurricanes). The dataset comprises more than 850,000 annotated masks of buildings.  Moreover, this dataset contains pre- and post-damage RGB images. In our study, we only focus on pre-damage images, therefore our natural hazard mappers had no access to information of weather hazards while learning the pre-task. For that reason, our selected pre-task and downstream tasks of natural hazard segmentation are not semantically entangled.

After pre-training of each natural hazard mapper on the pre-task data, we evaluate the generalization properties of the models on three downstream tasks when our methodology is applied. 
\newline 
\noindent  \textbf{Sen1Floods11 dataset:} is composed of 430 radar satellite images by Sentinel-1 capturing eleven globally-distributed flood events \cite{bonafilia2020sen1floods11}. The downstream task is the segmentation of the flood events in an unseen region based on two bands (i.e. VV and VH polarisation) and their multiplication (i.e. VV $\times$ VH).
\newline
\noindent  \textbf{Antarctic fracture detection dataset:} consists of 38 satellite images of six antarctic glaciers with fractures \cite{lai2020vulnerability}. In contrast to the data for the pre-task, this dataset is based on a single band (i.e., 125m-resolution MOA imagery from MODIS). We input the band to each channel of the model. The models need to segment glacier fractures in an unseen region.
\newline 
\noindent  \textbf{HR-GLDD dataset:} includes 1,758 satellite images depicting landslide from ten globally-distributed regions, where half of the events were rainfall triggered while the second half was earthquake-triggered \cite{meena2022hr}. The task of the models is to segment the extend of landslides in optical images (i.e. RGB images) of an unseen region.

We utilize the proposed train-test splits for each dataset (i.e., 90 test images for Sen1Floods11, 6 test images for Antarctic glacier dataset, and 354 test images for the HR-GLDD dataset). We report the results of our experiments on the test set. We did not make use of the train sets in the case of the zero-shot adaptation. For the unsupervised domain adaptation, we utilize a small number of images from the training dataset without access to their label.

\subsection{Natural hazard mapper training}
We utilized the training setting proposed in the xView2 challenge \cite{durnov2020xview}, including a weighted combination of focal and dice loss as our loss function for natural hazard mappers. All models were pre-trained over 50 epochs with early stopping by means of AdamW optimizer with initial learning rate of 0.00015 and a learning rate scheduler with gamma 0.5, and a batch size of 16. We applied a threshold to each of output of our models to obtain a classification of the pixels into the classes \textit{background} or \textit{natural hazard}. We utilized the 95\% percentile of the pixel distribution as threshold since this approach outperformed Otsu's thresholding in our initial experiments that are not reported in this paper. 

\begin{figure*}[ht]
    \centering
        \begin{subfigure}[b]{0.33\textwidth}
         \centering
         \includegraphics[width=1\textwidth]{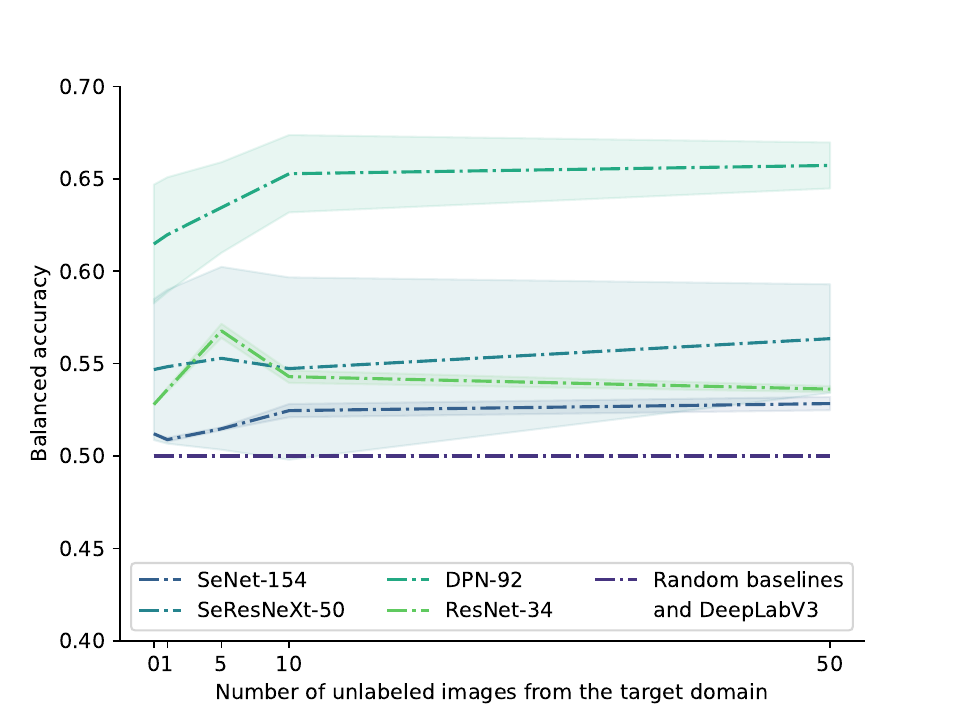}
         \caption{Sen1Floods11}
         \label{fig:sen1floods11}
     \end{subfigure}
     \begin{subfigure}[b]{0.33\textwidth}
          \centering
         \includegraphics[width=1\textwidth]{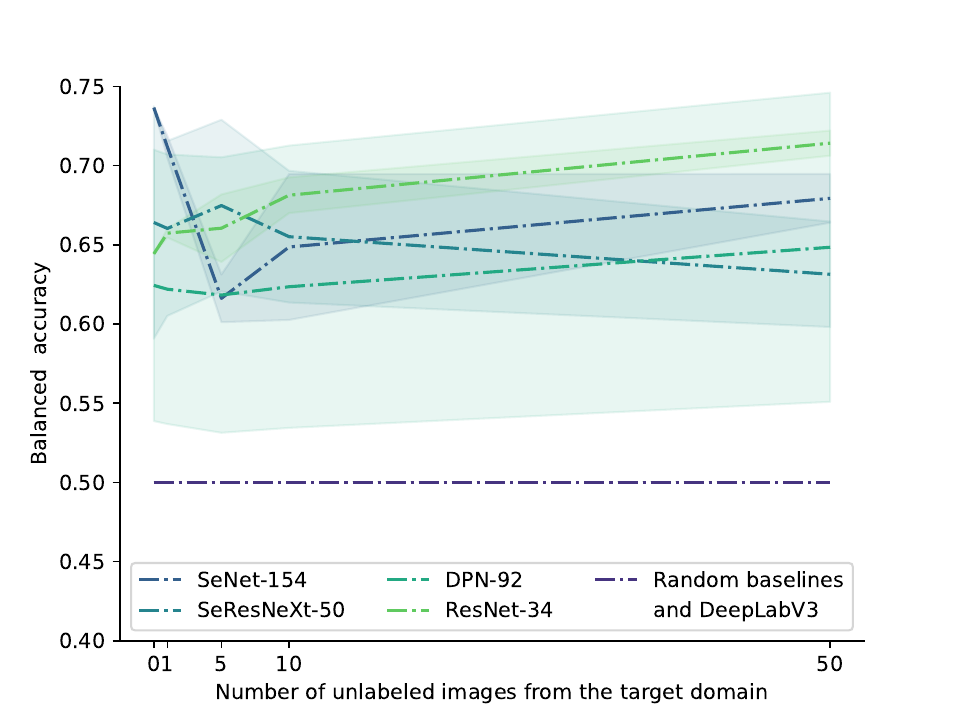}
         \caption{Antarctic Glaciers}
         \label{fig:antarctic}
     \end{subfigure}
     \begin{subfigure}[b]{0.33\textwidth}
         \centering
         \includegraphics[width=1\textwidth]{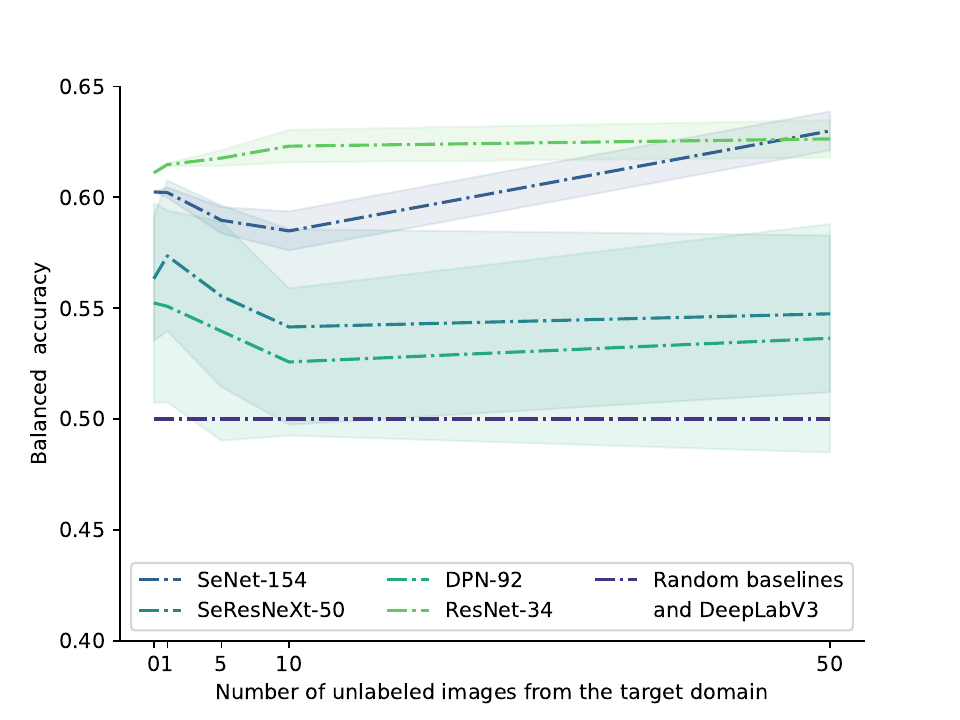}
         \caption{Landslide}
         \label{fig:landslide}
     \end{subfigure}
     \caption{Unsupervised generalization behavior of four different deep learning models pretrained on the localization of buildings in high-resolution satellite images and then tested on three downstream tasks: (a) flood mapping, (b) segmenting fractures in glaciers, and (c) landslide mapping. Graphs depict mean $\pm$ std of balanced accuracy.}
     \label{fig:eval-results}
\end{figure*}

\subsection{Evaluation protocol}
We evaluated the generalization of the natural hazard mappers without applying any domain adaptation technique (i.e. no fine-tuning) as our baseline. 
We benchmarked the performance of our models against two random baselines (i.e. uniformly-distributed noise and output generated by the selected U-Net architectures with random weights) and a state-of-the-art segmentation model, i.e. DeepLabV3. 
We then evaluated the generalization of natural hazard mappers when an unsupervised domain adaptation technique was applied. We made use of the available unlabeled data to update statistics of the batch normalization layers of our models and thereby adapted our models to the domain of each downstream task, accordingly. We evaluated different scenarios of unsupervised domain adaptation as part of $k$-shot adaptation (i.e., including $k \in \{1, 5, 10, 50\}$ unlabeled image(s) from the target domain to update batch normalization layers of natural hazard mappers).
For evaluation of semantic segmentation of three different natural hazards (i.e. pixel-wise classification), we calculated balanced accuracy, IoU and F1 Score. Since the two classes of natural hazard and background are highly unbalanced, we report balanced accuracy and IoU on the the class of interest.

\section{Results and discussion}

\begin{figure*}[ht]
    \centering
        \begin{subfigure}[b]{0.33\textwidth}
         \centering
         \includegraphics[width=0.765\textwidth]{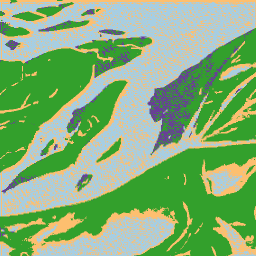}
         \caption{Sen1Floods11}
         \label{fig:sen1floods11_iou}
     \end{subfigure}
     \begin{subfigure}[b]{0.33\textwidth}
          \centering
         \includegraphics[width=0.765\textwidth]{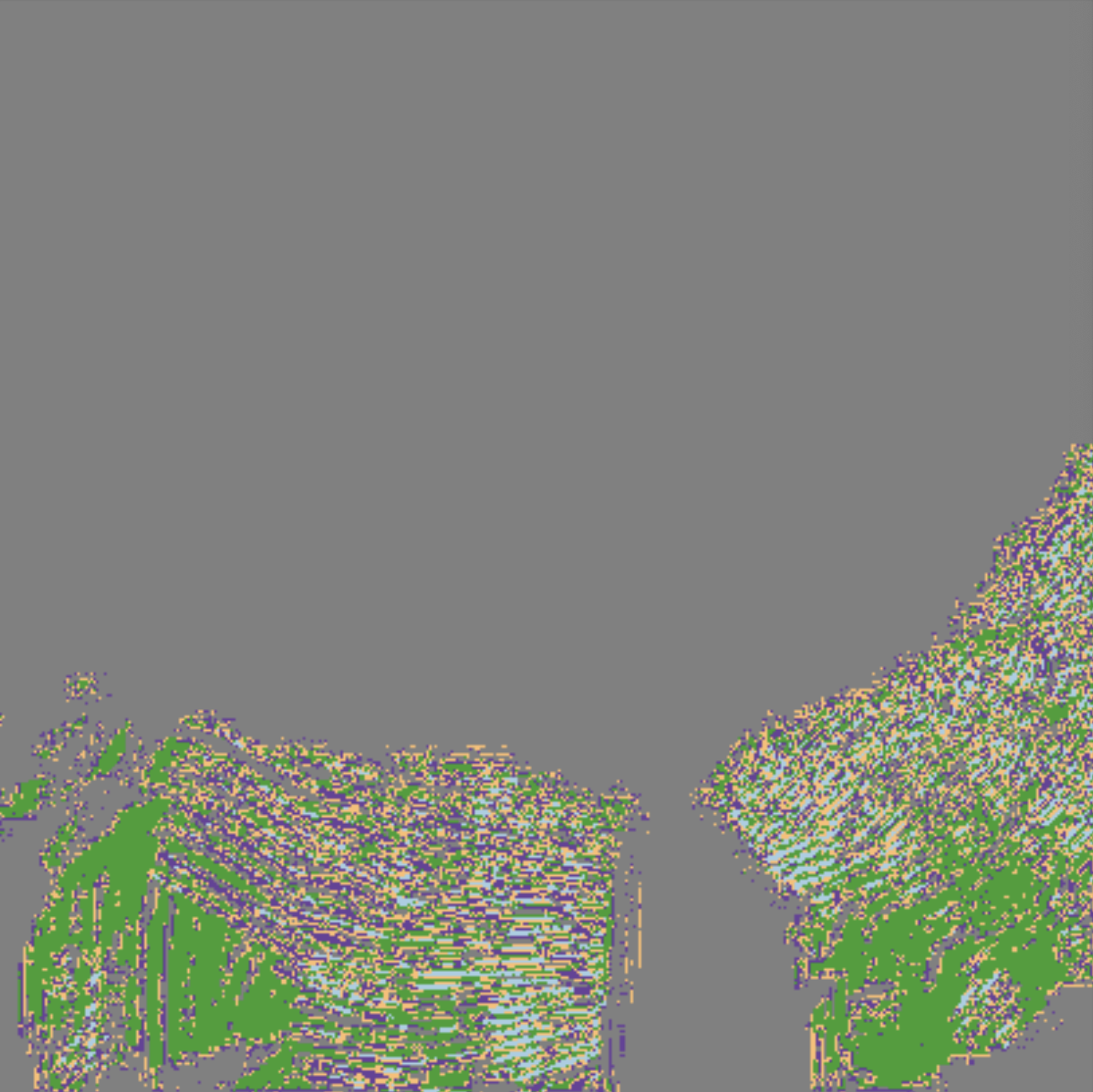}
         \caption{Antarctic Glaciers}
         \label{fig:antarctic_iou}
     \end{subfigure}
     \begin{subfigure}[b]{0.33\textwidth}
         \centering
         \includegraphics[width=0.765\textwidth]{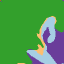}
         \caption{Landslide}
         \label{fig:landslide_iou}
     \end{subfigure}
     \caption{Zero-shot predictions depicting true negative (TN, \textcolor{Green}{$\bullet$}), true positive (TP, \textcolor{cyan}{$\bullet$}), false negative (FN, \textcolor{yellow}{$\bullet$}), and false positive (FP, \textcolor{violet}{$\bullet$}) pixel-level predictions using DPN-92 backbone. 
    Pixel that are not part of a glacier in Fig.~2(b) are visualized in \textcolor{gray}{$\bullet$}.
    }
     \label{fig:vis-results}
\end{figure*}

\subsection{Experimental results}
In Fig.~\ref{fig:eval-results}, we present results of our U-Net models, random baselines, and state-of-the-art DeepLabV3. Firstly, we find that the performance of DeepLabV3 pre-trained on the MS-COCO dataset does not significantly vary from random performance for the segmentation of natural hazards. Unsurprisingly, the model predicts each pixel to belong to the background class as none of the classes encountered in pre-training are part of the target domain. Thus, state-of-the-art models for image segmentation do not generalize to the segmentation of natural hazards sufficiently and require fine-tuning.
Secondly, without access to any data from the target domain and across different geo-locations, we confirm that performance of all four models is statistically significantly better than the baselines for the segmentation of landslide and fractures in glaciers confirmed by applying one-tailed Wilcoxon Rank tests with 0.01 significance level. For the segmentation of floods, two out of four models are statistically significantly better than the baselines. This emphasizes the importance of suitable pre-training for the generalization behavior of deep learning models for earth monitoring.
Thirdly, by utilizing few unlabeled images as part of the unsupervised domain adaptation, we observe a general trend in improving performance. For the majority of the models, an increase of the number of unlabeled images used for unsupervised domain adaptation leads to improvement of the model performance as reported in Table~\ref{tab:bal_acc} and Fig.~\ref{fig:eval-results}. 

\begin{table}[ht]
    \footnotesize
    \centering
 \begin{tabular}{l cccc}
        \toprule
              \textbf{\# images} & \textbf{ResNet-34} & \textbf{SeNet-154} & \textbf{SeResNeXt-50} & \textbf{DPN-92} \\
              \midrule
                \multicolumn{5}{c}{\textbf{Sen1Floods11}} \\
             \midrule
              0 & 0.5278 & 0.5120  & 0.5468***  & 0.6147***  \\
              \midrule
              1 & 0.5359  & 0.5089  &	0.5483  & 0.6197  \\
              5 & 0.5677  & 0.5147  &	0.5529  & 0.6345  \\
            \midrule
            \multicolumn{5}{c}{\textbf{Antarctic glaciers}} \\
              \midrule
              0 & 0.6443*** &	0.7366***  & 0.6641***  & 0.6244***  \\
              \midrule
              1 & 0.6573  &	0.7119  & 0.6603  & 0.6220  \\
              5 & 0.6605  &	0.6162  & 0.6748  & 0.6183 \\
            \midrule
            \multicolumn{5}{c}{\textbf{Landslide}} \\
              \midrule
              0 & 0.6110***  & 0.6024***  & 0.5633***  & 0.5523*** \\
              \midrule
              1 & 0.6147  & 0.6021  &	0.5736  & 0.5508  \\
              5 & 0.6177  & 0.5896  & 0.5554  & 0.5396  \\
              \bottomrule
             \end{tabular}
    \caption{Balanced accuracy scores of four deep learning models pretrained on the localization of buildings in high-resolution satellite images and then applied to three downstream tasks: (a) flood mapping, (b) segmenting fractures in glaciers, and (c) landslide mapping.}
    \label{tab:bal_acc}
\end{table}

\begin{table}[ht]
    \footnotesize
    \centering
 \begin{tabular}{l cccc}
        \toprule
              \textbf{\# images} & \textbf{ResNet-34} & \textbf{SeNet-154} & \textbf{SeResNeXt-50} & \textbf{DPN-92} \\
              \midrule
                \multicolumn{5}{c}{\textbf{Sen1Floods11 (class of interest)}} \\
             \midrule
              0 & 0.0305 & 0.0146 & 0.0292 & 0.1253 \\
              \midrule
              1 & 0.0370 & 0.0128 & 0.0270 & 0.1289 \\
              5 & 0.0643 & 0.0190 & 0.0325 & 0.1364 \\
            \midrule
            \multicolumn{5}{c}{\textbf{Antarctic glaciers (class of interest)}} \\
              \midrule
              0 & 0.0106 & 0.0156 & 0.0078 & 0.0166 \\
              \midrule
              1 & 0.0105 & 0.0144 & 0.0087 & 0.0165 \\
              5 & 0.0108 & 0.008 & 0.0093 & 0.0163 \\
            \midrule
            \multicolumn{5}{c}{\textbf{Landslide (class of interest)}} \\
              \midrule
              0 & 0.1270 & 0.1221 & 0.0618 & 0.1036\\
              \midrule
              1 &0.1281 & 0.1216 & 0.0841 & 0.1047 \\
              5 & 0.1314 & 0.1087 & 0.0822 & 0.0917 \\
              \bottomrule
             \end{tabular}
    \caption{IoU scores for the class of interest of four deep learning models pretrained on the localization of buildings in high-resolution satellite images and then applied to three downstream tasks: (a) flood mapping, (b) segmenting fractures in glaciers, and (c) landslide mapping.}
    \label{tab:iou_positive_class}
\end{table}

For the Sen1Floods11 dataset, we observe that the DPN-92 model achieves a balanced accuracy score in the pixel-wise segmentation of floods of 0.61 without access to any information from the target domain (i.e., zero-shot adaptation), as shown in Fig.~\ref{fig:sen1floods11} and Table~\ref{tab:bal_acc}. When adapting the model to the target domain, the performance increases up to 0.66 with access to 50 unlabeled images from the target domain. 
For the segmentation of fractures in glaciers shown in Fig.~\ref{fig:antarctic}, the ResNet-34 model achieves a score of 0.64 in the zero-shot setting (i.e. without access to unlabeled images from the target domain) that is increased up to a score of 0.71 in the 50-shot unsupervised domain adaptation (i.e. with access to 50 unlabeled images from the target domain). 
Finally, for the segmentation of landslide, the ResNet-34 model achieves a zero-shot performance of 0.61 that is improved up to 0.63 with access to 50 unlabeled images from the target domain as depicted in Fig.~\ref{fig:landslide}. 
Based on the unsupervised domain adaptation, the batch normalization layers and therefore also the masks produced by our models change. This also means that the distribution of pixels in the mask varies for different numbers of unlabeled images, which can cause non-optimal thresholding when utilizing the 95\% percentile. This explains individual performance drops with access to unlabeled images (e.g., 0-shot adaptation vs. 5-shot adaptation in Fig.~\ref{fig:antarctic}).

We observe similar patterns in zero-shot and few-shot unsupervised domain adaptation when evaluating the IoU scores on the positive class in Table~\ref{tab:iou_positive_class}. It is worth noting that the overall IoU scores are significantly higher than the scores on the positive class, ranging between 0.75 on Sen1Floods11 and 0.95 on Antarctic glaciers given the high class imbalance. 
Finally, we visualize an example prediction for each downstream task in Fig.~\ref{fig:vis-results}. We observe that our approach overall achieves a plausible performance on the segmentation of different kinds of natural hazards. However, there are regions where our method tends to overpredict (see Fig.~\ref{fig:vis-results}~(c)). 

\subsection{Limitations}
A priori selection of thresholds to distinguish natural hazards from the background may need to be adjusted during the unsupervised domain adaptation process due to varying characteristics of different geographic areas. Knowledge of changing distribution of geographic areas could be a key factor to make the proposed approach more accurate and precise. 
We anticipate that the usage of ancillary information, such as digital elevation model maps and land cover maps could be investigated to improve the current approach in the future. 

\section{Conclusion}
In a range of experiments, we demonstrate that a suitable pre-training improves the generalization of DL models for segmentation of unseen natural hazards.
Importantly, our approach is not limited to a specific modality (e.g., RGB or SAR channels) and allows for generalization of the models across different geographical regions and hazards. Our results prove that unsupervised domain adaptation can improve the performance of all considered models based on very few unlabeled samples.

\vfill

\pagebreak

\bibliographystyle{IEEEbib}
\bibliography{refs}

\begin{thebibliography}{10}

\bibitem{hallegatte2013future}
Stephane Hallegatte, Colin Green, Robert~J Nicholls, and Jan Corfee-Morlot,
\newblock ``Future flood losses in major coastal cities,''
\newblock {\em Nature Climate Change}, vol. 3, no. 9, pp. 802--806, 2013.

\bibitem{kainthura2022hybrid}
Poonam Kainthura and Neelam Sharma,
\newblock ``{Hybrid machine learning approach for landslide prediction,
  Uttarakhand, India},''
\newblock {\em Scientific Reports}, vol. 12, no. 1, pp. 1--23, 2022.

\bibitem{rounce2023global}
David~R. Rounce, Regine Hock, Fabien Maussion, Romain Hugonnet, William
  Kochtitzky, Matthias Huss, Etienne Berthier, Douglas Brinkerhoff, Loris
  Compagno, Luke Copland, Daniel Farinotti, Brian Menounos, and Robert~W.
  McNabb,
\newblock ``{Global glacier change in the 21st century: Every increase in
  temperature matters},''
\newblock {\em Science}, vol. 379, no. 6627, pp. 78--83, 2023.

\bibitem{ma2015remote}
Yan Ma, Haiping Wu, Lizhe Wang, Bormin Huang, Rajiv Ranjan, Albert Zomaya, and
  Wei Jie,
\newblock ``Remote sensing big data computing: Challenges and opportunities,''
\newblock {\em Future Generation Computer Systems}, vol. 51, pp. 47--60, 2015.

\bibitem{wang2022advancing}
Di~Wang, Qiming Zhang, Yufei Xu, Jing Zhang, Bo~Du, Dacheng Tao, and Liangpei
  Zhang,
\newblock ``Advancing plain vision transformer towards remote sensing
  foundation model,''
\newblock {\em IEEE Transactions on Geoscience and Remote Sensing}, 2022.

\bibitem{sun2022ringmo}
Xian Sun, Peijin Wang, Wanxuan Lu, Zicong Zhu, Xiaonan Lu, Qibin He, Junxi Li,
  Xuee Rong, Zhujun Yang, Hao Chang, et~al.,
\newblock ``{Ringmo: A remote sensing foundation model with masked image
  modeling},''
\newblock {\em IEEE Transactions on Geoscience and Remote Sensing}, 2022.

\bibitem{zhu2017deep}
Xiao~Xiang Zhu, Devis Tuia, Lichao Mou, Gui-Song Xia, Liangpei Zhang, Feng Xu,
  and Friedrich Fraundorfer,
\newblock ``{Deep learning in remote sensing: A comprehensive review and list
  of resources},''
\newblock {\em IEEE Geoscience and Remote Sensing Magazine}, vol. 5, no. 4, pp.
  8--36, 2017.

\bibitem{wang2022self}
Yi~Wang, Conrad~M Albrecht, Nassim Ait~Ali Braham, Lichao Mou, and Xiao~Xiang
  Zhu,
\newblock ``Self-supervised learning in remote sensing: A review,''
\newblock {\em arXiv preprint arXiv:2206.13188}, 2022.

\bibitem{zantedeschi2019cumulo}
Valentina Zantedeschi, Fabrizio Falasca, Alyson Douglas, Richard Strange,
  Matt~J Kusner, and Duncan Watson-Parris,
\newblock ``{Cumulo: A dataset for learning cloud classes},''
\newblock {\em arXiv preprint arXiv:1911.04227}, 2019.

\bibitem{wang2020weakly}
Sherrie Wang, William Chen, Sang~Michael Xie, George Azzari, and David~B
  Lobell,
\newblock ``Weakly supervised deep learning for segmentation of remote sensing
  imagery,''
\newblock {\em Remote Sensing}, vol. 12, no. 2, pp. 207, 2020.

\bibitem{fraccaro2022deploying}
Paolo Fraccaro, Nikola Stoyanov, Zaheed Gaffoor, Laura Elena~Cue La~Rosa,
  Jitendra Singh, Tatsuya Ishikawa, Blair Edwards, Anne Jones, and Komminist
  Weldermariam,
\newblock ``{Deploying an artificial intelligence application to detect flood
  from Sentinel 1 data},''
\newblock in {\em Proceedings of the Thirty-Sixth AAAI Conference on Artificial
  Intelligence}, 2022.

\bibitem{he2016deep}
Kaiming He, Xiangyu Zhang, Shaoqing Ren, and Jian Sun,
\newblock ``Deep residual learning for image recognition,''
\newblock in {\em Proceedings of the IEEE Conference on Computer Vision and
  Pattern Recognition}, 2016, pp. 770--778.

\bibitem{hu2018squeeze}
Jie Hu, Li~Shen, and Gang Sun,
\newblock ``Squeeze-and-excitation networks,''
\newblock in {\em Proceedings of the IEEE Conference on Computer Vision and
  Pattern Recognition}, 2018, pp. 7132--7141.

\bibitem{xie2017aggregated}
Saining Xie, Ross Girshick, Piotr Doll{\'a}r, Zhuowen Tu, and Kaiming He,
\newblock ``Aggregated residual transformations for deep neural networks,''
\newblock in {\em Proceedings of the IEEE Conference on Computer Vision and
  Pattern Recognition}, 2017, pp. 1492--1500.

\bibitem{chen2017dual}
Yunpeng Chen, Jianan Li, Huaxin Xiao, Xiaojie Jin, Shuicheng Yan, and Jiashi
  Feng,
\newblock ``Dual path networks,''
\newblock {\em Advances in Neural Information Processing Systems}, vol. 30,
  2017.

\bibitem{gupta2019xbd}
Ritwik Gupta, Richard Hosfelt, Sandra Sajeev, Nirav Patel, Bryce Goodman, Jigar
  Doshi, Eric Heim, Howie Choset, and Matthew Gaston,
\newblock ``{xBD: A dataset for assessing building damage from satellite
  imagery},''
\newblock {\em arXiv preprint arXiv:1911.09296}, 2019.

\bibitem{bonafilia2020sen1floods11}
Derrick Bonafilia, Beth Tellman, Tyler Anderson, and Erica Issenberg,
\newblock ``{Sen1Floods11: A georeferenced dataset to train and test deep
  learning flood algorithms for Sentinel-1},''
\newblock in {\em Proceedings of the IEEE/CVF Conference on Computer Vision and
  Pattern Recognition Workshops}, 2020, pp. 210--211.

\bibitem{lai2020vulnerability}
Ching-Yao Lai, Jonathan Kingslake, Martin~G Wearing, Po-Hsuan~Cameron Chen,
  Pierre Gentine, Harold Li, Julian~J Spergel, and J~Melchior van Wessem,
\newblock ``{Vulnerability of Antarctica’s ice shelves to meltwater-driven
  fracture},''
\newblock {\em Nature}, vol. 584, no. 7822, pp. 574--578, 2020.

\bibitem{meena2022hr}
Sansar~Raj Meena, Lorenzo Nava, Kushanav Bhuyan, Silvia Puliero, Lucas~Pedrosa
  Soares, Helen~Cristina Dias, Mario Floris, and Filippo Catani,
\newblock ``{HR-GLDD: A globally distributed dataset using generalized DL for
  rapid landslide mapping on HR satellite imagery},''
\newblock {\em Earth System Science Data Discussions}, pp. 1--21, 2022.

\bibitem{durnov2020xview}
Victor Durnov,
\newblock ``{xView2: 1st place solution},'' GitHub,
  https://github.com/vdurnov/xview2\_1st\_place\_solution, 2020.

\end{thebibliography}

\end{document}